\documentclass{article} % For LaTeX2e
\usepackage{iclr2024_conference,times}

\usepackage{amsmath,amsfonts,bm}
\usepackage{wrapfig}

\def\eqref#1{equation~\ref{#1}}
\def\1{\bm{1}}

\DeclareMathAlphabet{\mathsfit}{\encodingdefault}{\sfdefault}{m}{sl}
\SetMathAlphabet{\mathsfit}{bold}{\encodingdefault}{\sfdefault}{bx}{n}

\usepackage{hyperref}
\usepackage{url}
\usepackage[utf8]{inputenc} % allow utf-8 input
\usepackage[T1]{fontenc}    % use 8-bit T1 fonts
\usepackage{booktabs}       % professional-quality tables
\usepackage{amsfonts}       % blackboard math symbols
\usepackage{nicefrac}       % compact symbols for 1/2, etc.
\usepackage{microtype}      % microtypography
\usepackage{xcolor}  
\usepackage{subfigure}
\usepackage{multirow}
\usepackage{graphicx}
\usepackage{makecell}

\newcommand{\squishlist}{
\begin{list}{$\bullet$}
{ \usecounter{Lcount}
\setlength{\itemsep}{0pt}
\setlength{\parsep}{0pt}
\setlength{\topsep}{0pt}
\setlength{\partopsep}{0pt}
\setlength{\leftmargin}{2em}
\setlength{\labelwidth}{1.5em}
\setlength{\labelsep}{0.5em} } }
\newcommand{\squishend}{
\end{list} }

\title{Do Large Language Models Know about Facts?}

\author{
Xuming Hu\textsuperscript{1}, Junzhe Chen\textsuperscript{1}, Xiaochuan Li\textsuperscript{1}, \bf{Yufei Guo}\textsuperscript{1}, \bf{Lijie Wen}\textsuperscript{1}, \\~\bf{Philip S. Yu}\textsuperscript{2}, \bf{Zhijiang Guo}\textsuperscript{3}\\
~\textsuperscript{1} Tsinghua University~~~
\textsuperscript{2} University of Illinois at Chicago~~~\\
~\textsuperscript{3} University of Cambridge~~~\\
~{\tt\small hxm19@mails.tsinghua.edu.cn, zg283@cam.ac.uk}
}

\iclrfinalcopy % Uncomment for camera-ready version, but NOT for submission.
\begin{document}

\maketitle

% \section{Submission of conference papers to ICLR 2024}

\begin{abstract}
% Recently, large language models (LLMs) have driven striking performance improvements across a range of natural language processing tasks. Despite their capabilities, LLMs have similar limitations as earlier natural language generation models. Most importantly, they are still not fully reliable, as they sometimes hallucinate facts and perform untruthful reasoning. It is not yet known whether LLMs encode factually accurate information. To this end, we aim to comprehensively evaluate the extent and scope of factual knowledge within LLMs by designing the benchmark Pinocchio. Pinocchio contains 20K diverse factual questions that span different sources, timelines, domains, regions, and languages. Furthermore, we investigate whether LLMs are able to compose multiple facts, update factual knowledge temporally, reason over multiple facts, identify subtle factual differences and resist adversarial examples. Extensive experiments on different sizes and types of LLMs show that existing LLMs still lack factual knowledge and suffer from various spurious correlations. We believe this is a critical bottleneck for realizing trustworthy artificial intelligence.  Pinocchio and our codes will be publicly available.

Large language models (LLMs) have recently driven striking performance improvements across a range of natural language processing tasks. The factual knowledge acquired during pretraining and instruction tuning can be useful in various downstream tasks, such as question answering, and language generation. Unlike conventional Knowledge Bases (KBs) that explicitly store factual knowledge, LLMs implicitly store facts in their parameters. Content generated by the LLMs can often exhibit inaccuracies or deviations from the truth, due to facts that can be incorrectly induced or become obsolete over time. To this end, we aim to comprehensively evaluate the extent and scope of factual knowledge within LLMs by designing the benchmark Pinocchio. Pinocchio contains 20K diverse factual questions that span different sources, timelines, domains, regions, and languages. Furthermore, we investigate whether LLMs are able to compose multiple facts, update factual knowledge temporally, reason over multiple pieces of facts, identify subtle factual differences, and resist adversarial examples. Extensive experiments on different sizes and types of LLMs show that existing LLMs still lack factual knowledge and suffer from various spurious correlations. We believe this is a critical bottleneck for realizing trustworthy artificial intelligence. The dataset Pinocchio and our codes will be publicly available.

\end{abstract}
\section{Introduction}

Large language models (LLMs) have revolutionized natural language processing (NLP) in recent years since they have significantly improved performance on various downstream tasks~\citep{BrownMRSKDNSSAA20, Chowdhery2022, Ouyang0JAWMZASR22, touvron2023, llama2,chatgpt2022, gpt4}. 
Prior efforts have shown that language models can store factual knowledge and act as knowledge bases~\citep{PetroniRRLBWM19, JiangXAN20}. Factual knowledge in language models acquired during pretraining can benefit knowledge-intensive downstream tasks such as question answering and fact checking~\citep{RobertsRS20,wenhao2023,PanWLLWKN23}.

Despite advancements in LLMs, they still struggle with generating content that exhibits inaccuracies or deviations from the facts and making reasoning errors~\citep{LinHE22,bubeck2023}. These factual errors can be difficult to identify since LLMs implicitly memorize facts through their parameters rather than explicitly store factual knowledge as traditional Knowledge Bases. Accessing and interpreting the computations and memories of these models can be challenging~\citep{RibeiroSG16a,BelinkovG19}, especially when APIs are the only means of interaction and many interpretation methods rely on weights and representations~\citep{CaoSAT20}.  The presence of errors in stored factual knowledge or the incorrect induction and obsolescence of certain facts over time may be contributing factors to this limitation, which in turn affects the performance of LLMs~\citep{ElazarKRRHSG21,CaoLHSYLXX20}. This limitation restricts the application of LLMs in some high-stakes areas, such as healthcare, finance, and law~\citep{DongDSXSL22}. Hence, exploring the degree to which LLMs hold factual information and their ability to reason with such knowledge is vital.

% Content that is automatically generated can often exhibit inaccuracies or deviations from the facts due to the limited capacity of large language models (LLMs). LLMs are susceptible to producing content that appears credible but may actually be factually incorrect or imprecise.

% leading to deception and decreased trust.

\begin{figure}
  \centering
  \includegraphics[scale=0.75]{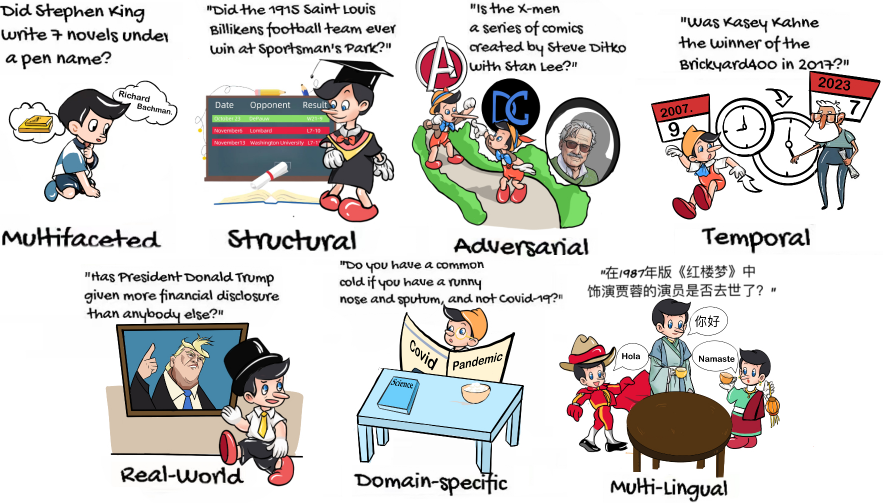}
  \vspace{-4mm}
  \caption{Pinocchio is a comprehensive dataset that tackles 7 distinct tasks related to factual knowledge and reasoning. It consists of 20,713 multiple-choice questions that have been sourced from various reliable and diverse channels.}
  \label{figure:Pinocchios}
  \vspace{-5mm}
\end{figure}

% This limitation restricts the application of generative AI in some high-stakes areas, such as healthcare, finance, and law. Therefore, it is crucial to identify these errors systematically to improve the usefulness and reliability of the generated content.

To this end, we propose the Pinocchio benchmark, a comprehensive testbed of factuality and reasoning designed for LLMs. Pinocchio contains 20K diverse factual questions that span different sources, timelines, domains, regions, and languages. Furthermore, we investigate whether LLMs are able to recognize the combination of multiple facts, reason over structured and unstructured evidence, realize facts change over time, identify subtle factual differences, and resist adversarial examples. We control for problem difficulty in each distinct reasoning task to enable fine-grained analysis. 

With the Pinocchio benchmark, we evaluate whether various LLMs~\citep{scao2022bloom,OPT2022,Ouyang0JAWMZASR22,flan2022,touvron2023,vicuna2023} could store factual knowledge and perform reasoning based on it. We envision Pinocchio as a suite of benchmarks, subsets of which could be separately utilized to assess certain model abilities of interest and analyze important strengths and limitations of LLMs. For instance, in temporal tasks, we find that LLMs lack factual knowledge for up-to-date questions; in complex factual tasks that require multi-hop reasoning, LLMs still have limitations, even when various prompting strategies are employed. We hope Pinocchio can guide the researchers to understand the abilities of their models from multiple dimensions and facilitate the development of factual knowledge in LLMs.

\section{Dataset Construction}

\subsection{Tasks}

% In this work, we hope to explore whether large language models (LLMs) have the ability to know facts. Table~\ref{dataset-size} summarize the proposed dataset. Specifically, we will comprehensively study it from the following aspects: 

Aiming to systematically evaluate the factual knowledge and related reasoning abilities of LLMs, we raise seven research questions, then carefully select factual statements from different sources summarized in Table~\ref{dataset-size}.

\squishlist
    \item \textbf{Task 1: Multifacted} 
    Previous research~\citep{PetroniRRLBWM19} has shown that small language models like BERT have the ability to retain relational knowledge from training data and answer ``fill-in-the-blank'' cloze statements. This raises the question of \textit{whether LLMs can also store and reason over multiple pieces of facts obtained during pretraining}. It is not just important for LLMs to memorize individual facts accurately, but to also recognize and generate new combinations of facts from different sources. To investigate this issue, we have selected claims from the FEVER dataset~\citep{ThorneVCM18}, which were written by human annotators based on information from Wikipedia articles. These claims are either supported or refuted by multiple facts from (the same or several) Wikipedia articles, or there is insufficient information available to verify them. To assess the performance of language models in handling various combinations of facts, we have sampled statements that require different numbers of evidence, ranging from one to many, enabling fine-grained analysis.
    \item \textbf{Task 2: Structural} In addition to unstructured text, factual knowledge is also commonly stored in a structured format, such as tables, lists, or databases~\citep{BhagavatulaND13}. However, current LLMs are primarily trained on unstructured text using next word prediction loss~\citep{BrownMRSKDNSSAA20,touvron2023}. In order to process structured data, it is often converted into text strings using various methods, such as linearizing tables. This raises the question of \textit{whether LLMs are capable of effectively memorizing and reasoning over facts from structured sources, similar to their performance with unstructured text}. To investigate this question, we sample factual statements from the FEVEROUS dataset~\citep{AlyGST00CM21}, which is constructed in a similar manner to FEVER but includes evidence in the form of tables, sentences, or both.
    \item \textbf{Task 3: Adversarial} 
    Language models are known to be vulnerable to adversarial examples that are strategically modified to deceive even advanced models with hardly noticeable changes~\citep{Shen2023}. Given this knowledge, it is important to examine \textit{whether LLMs can withstand adversarial examples in the context of factuality}. To investigate this, we utilize two datasets, namely Symmetric~\citep{SchusterSYFSB19} and FM2~\citep{EisenschlosDBBB21}. These datasets consist of adversarial examples that have been crafted using various strategies, including temporal inference and diverting to unrelated facts. 
    \item \textbf{Task 4: Temporal} 
    Facts are not static but rather possess a dynamic nature. With the vast amount of new information constantly emerging, facts often undergo changes, additions, or alterations. It raises the question of \textit{whether LLMs are able to adapt to these factual changes over time}. In particular, we wonder if LLMs are capable of discerning factual knowledge from different time periods, since the pretraining corpus may not be processed and organized chronologically. To explore this, we utilize the VitaminC~\citep{schuster-etal-2021-get} dataset, which consists of claims based on modifications made to factual content in Wikipedia articles. Claims can be either refuted by outdated facts or supported by updated facts.
    \item \textbf{Task 5: Real-World} 
    In contrast to other tasks that assume Wikipedia has all the essential factual information, verifying viral claims on the internet often requires not only factual knowledge from various sources but also common sense and worldly knowledge. An important query we have is \textit{whether LLMs can effectively integrate diverse types and sources of knowledge acquired during training}. To address this, we select claims from the FactCheck~\citep{Poltifact2022} dataset, which consists of claims spread over the Internet and subsequently verified by journalists.
    \item \textbf{Task 6: Domain-Specific} 
    In addition to the tasks mentioned earlier, which primarily focus on factual knowledge in general domains, we are also interested in exploring \textit{how LLMs possess the capability to access domain-specific factual knowledge}. The domain-specific setting presents unique challenges. Take the science domain as an example, LLMs need to acquire background knowledge, handle quantitative reasoning, and comprehend specialized statistical language. To investigate this further, we sample claims from PubHealth \citep{kotonya-toni-2020-explainable-automated} in the public health domain and SciFact \citep{wadden-etal-2022-scifact} in the science domain. 
    \item \textbf{Task 7: Multi-Lingual} 
     Existing LLMs are mainly trained on English corpus because of their abundance and quality~\citep{Chowdhery2022,touvron2023}. However, the scarcity of training data in other languages raises the question of \textit{whether LLMs can transfer the factual knowledge acquired in English to other languages}. To investigate this, we collected claims from various languages including French, Chinese, and more, using the XFACT dataset~\citep{gupta-srikumar-2021-x} and the CHEF dataset~\citep{hu-etal-2022-chef} in a total of 27 different languages.
\squishend

\begin{table}[t!]
  \caption{Pinocchio Dataset Sources, Descriptions, and Data Distribution.}
  \label{dataset-size}
  \centering
   \scalebox{0.74}{
  \begin{tabular}{ccccccc}
    \toprule
    \multirow{2}{*}{Domain}  & \multirow{2}{*}{Description}  &  \multirow{2}{*}{Sources}& \multicolumn{4}{c}{Distribution} \\
    \cmidrule(r){4-7}
    &&&Fact. & Non-Fact. & NEI & ALL \\
    \midrule
    Multifaceted & Contain multiple facts & FEVER &1,111 &1,111 & 1,110 &3,332 \\
    Structural &  Contain structured and unstructured facts  & FEVEROUS &1,741 &1,953 &250  &3,944  \\ 
    Adversarial & Contain facts edited by adversarial methods & Symmetric, FM2 &815  &921  &- &1,736 \\
    Temporal & Contain facts that change over time & VitaminC &1,898 &1,043 &355 &3,296  \\
    Real-World & Contain factual statements spread online & PolitiFact &986  &1,987 &609 &3,582 \\
    Domain-Specific & Contain facts from health and science domains  & PubHealth, SciFact & 1,156 &715 &737 &2,608 \\
    Multi-Lingual & Contain facts in different languages & XFact, CHEF & 820 &848 &547 &2,215 \\
    % All&& &8,713 &9,619 &3,608 &21,940\\
    \bottomrule
  \end{tabular}}
  \vspace{-6mm}
% \label{tab:main}

\end{table}

% FEVER~\citep{ThorneVCM18}, FEVEROUS~\citep{AlyGST00CM21}, FoolMeTwice~\citep{EisenschlosDBBB21}, SciFact~\citep{wadden-etal-2022-scifact}, XFact~\citep{gupta-srikumar-2021-x}, PolitiFact, VitaminC~\citep{schuster-etal-2021-get}, and Symmetric~\citep{SchusterSYFSB19}.

\subsection{Annotation and Quality Control}

% When sampling from existing fact checking datasets, we only need the claims and corresponding labels from the dataset; the evidence required to verify the claims should come from the LLMs. The claims in existing datasets are declarative sentences, so we need to manually modify them into interrogative sentences with the same factuality as the original claims. 
We only select questions of a multi-choice format, similar to other benchmarks~\citep{HendrycksBKABTS21,Zhong2023}, because metrics are clearly defined (i.e. accuracy), and multi-choice questions are a simple but good proxy to evaluate the potential of advanced abilities of LLMs, which we consider could be easily exploited and reflected in various downstream applications through specialized instruction tuning. Each question has four choices and only one choice is the correct answer. LLMs are intended to be used to solve these questions through prompting. 

Specifically, we hired 10 undergraduate students, all with good English proficiency. We asked the students to rewrite the original claims into questions without distorting factuality while providing factuality labels for the questions. By transforming declarative statements into questions, using a Question-Answering approach can more effectively elicit factual knowledge from LLMs~\citep{Kadavath2022,LinHE22}, and we also illustrate through experiments in Sec. \ref{prompt_strategy_analysis}. Note that claims in the original datasets are usually labeled based on given evidence, e.g. evidence supports or refutes the claim, but in Pinocchio, we only need to judge the factuality of the question. So we use unified labels: Yes, No, Not Sure Enough. The three labels correspond respectively to Factual, Non-Factual, and Not Enough Information for factual questions. Considering that all fact-checking datasets use a three-label system~\citep{guo-etal-2022-survey}, we did not modify the number of labels to maintain consistency in labeling. When dealing with factuality questions in low-resource languages, for Chinese, the 5 undergraduate students we hired are native Chinese speakers. For other low-resource languages, we first use Google Translate to translate them into English and generate factuality questions, then translate the English questions back to the corresponding languages. The label distribution is shown in Table~\ref{dataset-size}. We paid the annotators accordingly based on the quantity and quality of the annotations.

We ensure the quality of the annotated factuality questions in two ways. The two authors of this paper served as meta-reviewers, sampling 10 questions from each of the three categories across the seven domains in Pinocchio. The meta-reviewers judged if the factuality labels were correct. For the 210 factuality questions, the average label accuracy was 92.4\%. We divided the 10 students into two groups and had each group re-annotate a random 200 questions annotated by the other group, then calculated inter-annotator agreement (IAA). The final IAA was 85.6\%. Based on meta-reviewer results and IAA, the factuality labels in Pinocchio are of good quality.

\section{Methodology}

\begin{figure}
  \centering
  \includegraphics[scale=0.78]{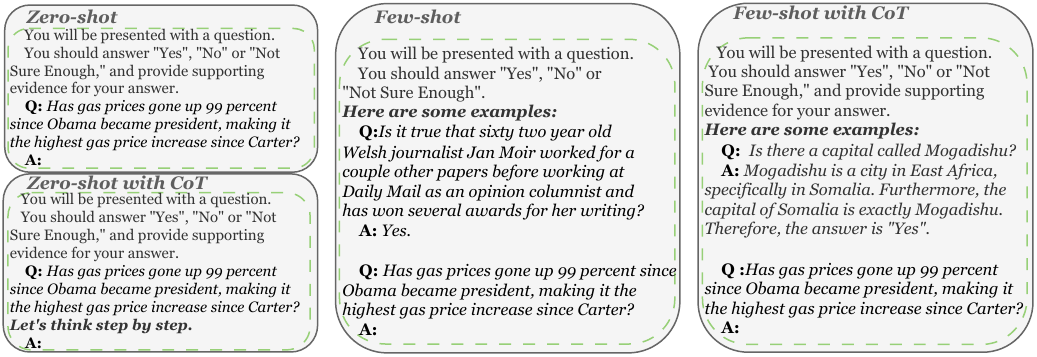}
  \vspace{-3mm}
  \caption{Illustration of prompts using different settings.}
  \label{figure:prompt}
   \vspace{-7mm}
\end{figure}

\subsection{Models}

To give a comprehensive view of the status of large language models in a factual context, we evaluate 10 accessible LLMs, undergone different training stages including pretraining, instruction tuning, and reinforcement learning from human feedback~\citep{Ouyang0JAWMZASR22}, covering diverse organizations and varying in size, as shown in Table~\ref{main-result}.

For pretraining LLMs, we adopt OPT~\citep{OPT2022}, BLOOM~\citep{Bloom2022}, and LLaMA~\citep{touvron2023}. For instruction-tuned LLMs, we adopt Alpaca~\citep{alpaca2023}, Vicuna~\citep{vicuna2023}, Flan
-T5~\citep{flan2022}, and ChatGLM~\citep{ZengLDWL0YXZXTM23}. After undergoing pretraining, instruction tuning, and RLHF, ChatGPT~\citep{chatgpt2022} is also taken into consideration. A detailed description of these models can be found in Appendix \ref{LLM_INTRO}.

\subsection{Prompt Strategy}

As illustrated in Figure~\ref{figure:prompt}, we employ 4 types of prompts to elicit desired responses from LLMs, namely: Zero-shot, Zero-shot with CoT~\citep{KojimaGRMI22}, Few-shot, and Few-shot with CoT~\citep{Wei0SBIXCLZ22}. Specifically, we begin by providing the model with task instruction, denoted as $Z$: ``You will be given a question. You should answer whether it is Yes, No, or Not Sure Enough and show your evidence''. This instruction informs the LLMs about the expected input and output. Subsequently, for any given input $Q$, we anticipate obtaining an output label $Y$ from the LLMs $f$: $Y=f(Q,Z)$. 

% We then provide detailed descriptions of these four variations of $Q$.

\paragraph{Zero-Shot Prompt}

In the zero-shot setting, the LLMs are expected to provide answers based on the Question $Q$ and the task instruction $Z$. We anticipate that the LLMs can directly generate the factual answer ``No'' when presented with $Q$: ``Has gas prices gone up 99 percent since Obama became president, making it the highest gas price increase since Carter?'' The zero-shot with CoT setting extends the question $Q$ by adding a two-stage prompt \citep{KojimaGRMI22}: ``Let's think step by step'', designed to encourage the LLMs to contemplate the process of determining the factual label $Y$.

\paragraph{Few-Shot Prompt}
In the few-shot setting, we employ three prompts: Yes, No, and Not Sure Enough, as query questions ($Q$) for model input. Due to space constraints, detailed examples of the prompts in Figure \ref{figure:prompt} are presented in Appendix \ref{Complete_prompts}. The utilization of the few-shot setup allows us to better tap into the inherent factual reasoning abilities of LLMs \citep{flan2022,chatgpt2022,wang2022self,alpaca2023}. This is particularly advantageous for those models that have not been fine-tuned with specific instructions, as their factual reasoning capabilities can be showcased through effective few-shot guidance. In contrast, they may struggle to adhere to instructions in zero-shot evaluations, thereby impacting their ultimate factual performance.

In the few-shot with CoT setting, we provide potential reasoning instructions to the LLMs before presenting the factual label ($Y$). The aim is to elicit the LLMs' innate factual reasoning abilities through examples of reasoning. As shown in Figure \ref{figure:prompt}, for the $Q$: ``Is there a capital called Mogadish?'' Our reasoning approach entails first explaining the noun phrase in the $Q$ (the subject and object), and subsequently elaborating on modifying phrases such as predicates or adjectives. Regarding the subject ``Mogadish'', we begin by furnishing a detailed definition: ``Mogadishu is a city in East Africa, specifically in Somalia.'' Following this, we proceed to reason about the relation between ``Mogadish'' and ``capital'': ``Furthermore, the capital of Somalia is indeed Mogadishu.'' Consequently, we arrive at the ultimate factual label: ``Therefore, the answer is Yes.'' We anticipate that the reasoning instructions provided to the LLMs will serve to stimulate its factual reasoning abilities.

\section{Experiments}
In the previous sections, we provided a detailed description of how Pinocchio was constructed and the LLMs used. In this section, we will begin by introducing the performance of various LLMs on Pinocchio across different settings and tasks, along with a detailed analysis.

\subsection{Main Results}
In Table \ref{main-result}, we present the average results of 10 accessible LLMs operating under varying settings on Pinocchio, run three times each. From Table \ref{main-result}, we draw the following conclusions:

\begin{table}[t!]
  \caption{Results obtained using different forms of prompts on 10 accessible LLMs.}
  \label{main-result}
  \centering
    \scalebox{0.74}{
\begin{tabular}{ccccccccccc}
\toprule
\multirow{2}{*}{Methods} & \multicolumn{2}{c}{Zero-shot w/o CoT} & \multicolumn{2}{c}{Zero-shot w/ CoT} & \multicolumn{2}{c}{Few-shot w/o CoT} & \multicolumn{2}{c}{Few-shot w/ CoT} & \multicolumn{2}{c}{Overall Performance} \\ \cmidrule(lr){2-3} \cmidrule(lr){4-5} \cmidrule(lr){6-7}  \cmidrule(lr){8-9} \cmidrule(lr){10-11}
                         & Accuracy          & F1                & Accuracy          & F1               & Accuracy          & F1               & Accuracy         & F1               & Accuracy           & F1                 \\ \midrule
OPT-6.7B                 & —               & —               & —               & —              & 36.9              & 27.9             & 37.9             & 28.5             & 18.8               & 14.3               \\
BLOOM-7B                 & 29.7              & 26.2              & 14.8              & 18.1             & 29.7              & 28.1             & 6.6              & 12.2             & 20.2               & 21.2               \\
LLaMA-7B                 & 31.8              & 29.6              & 22.3              & 24.9             & 36.8              & 28.6             & 35.3             & 31.4             & 31.6               & 28.6               \\ \midrule \midrule
Alpaca-7B                & 40.2              & 23.7              & 33.7              & 24.4             & 37.9              & 24.9             & 39.4             & 26.2             & 37.8               & 24.8               \\
Vicuna-7B                & 33.2              & 33.6              & 34.2              & 32.9             & 35.5              & 34.8             & 48.5             & 40.6             & 37.9               & 34.9               \\
Vicuna-13B               & 42.6              & 35.6              & 44.0              & 36.9             & 47.0              & 38.6             & \underline {47.0}       & 42.5             & 45.2               & 38.4               \\
ChatGLM-6B               & 37.4              & 31.0              & 36.5              & 31.7             & 41.6              & 37.9             & 42.9             & 37.5             & 39.6               & 34.5               \\
Flan-T5-11B              & 24.6              & 21.5              & 29.9              & 29.3             & 25.9              & 23.7             & 38.4             & 38.4             & 29.7               & 26.9               \\ \midrule \midrule
Text-Davinci-002         & {\underline {45.2}}        & 36.2              & {\underline{45.7}}        & 37.3             & 46.6              & 40.4             & 46.2             & 42.5             & {\underline {45.9}}         & 39.1               \\
Text-Davinci-003         & 42.8              & {\underline {41.4}}        & 43.1              & {\underline {42.1}}       & \textbf{48.8}     & {\underline {43.2}}       & 46.9             & {\underline {43.4}}       & 45.5               & {\underline {42.5}}         \\
GPT-3.5-Turbo            & \textbf{46.9}     & \textbf{44.3}     & \textbf{46.8}     & \textbf{44.4}    & {\underline {47.2}}        & \textbf{44.7}    & \textbf{47.1}    & \textbf{45.7}    & \textbf{47.0}      & \textbf{44.8}      \\ 
% \midrule
% Random Guess             & —                 & —                 & —                 & —                & —                 & —                & —                & —                & 33.3               & 29.8               \\ 
\bottomrule
\end{tabular}
}
\vspace{-5mm} 
\end{table}
\squishlist
    \item Regarding overall performance, we observe that, on average, LLMs without instruction tuning underperform those with instruction tuning by 16.0\%. GPT family LLMs undergoing RLHF exhibit superior results, indicating that instruction tuning and RLHF optimize alignment with human knowledge, thereby improving factual question response accuracy.
    \item Results obtained using the Few-shot setting significantly outperform those obtained when simply asking factual questions to LLMs in the Zero-shot setting, especially for models without RLHF, exhibiting an average improvement of 7.3\%. This highlights the capability of some sample prompts to better extract the inherent factual knowledge of LLMs.
    \item Using the CoT method, we observed a relative boost in performance in LLMs subjected to instruction tuning and RLHF, improving by an average of 2.1\%. Notably, the factual accuracy of LLMs like OPT, BLOOM, and LLaMA was mostly stable or even decreased. A review of outputs from these untuned LLMs revealed that, post-CoT application, LLMs tend to produce related content considerations, and extensive considerations often overshadow factual discernment tasks, causing incorrect factual label outputs. In contrast, for instruction-tuned LLMs, the CoT method facilitates enhanced exploration of factual entity relations in questions, resulting in accurate factual labels. See Appendix \ref{case-study} for detailed case analyses.
    \item The OPT model, without being tuned to instructions, struggles significantly to output correct factual labels under the settings of Zero-shot and Zero-shot CoT, often resulting in either a repetition of the original question or a refusal to output any content at all. This issue is somewhat alleviated under the settings of Few-shot and Few-shot CoT.
    \item Additionally, we studied the hyperparameters of LLMs. Due to limited computing resources, we only explored Vicuna-7B and Vicuna-13B. We found that as model parameters increase, performance on factual questions improves correspondingly, with an average increase of 5.4\%. This indicates that LLMs with more parameters can store more world knowledge and have stronger factual knowledge recognition capabilities.
    % \item Considering the imbalances in factual label categories within Pinocchio, we demonstrate the outcomes when a model randomly assigns a factual label. The results of random guessing fall between LLMs with and without instruction tuning, indicating that there is still significant room for improvement in the performance of LLMs on factual questions.
\squishend

In Table \ref{domain-result}, we present the factual performance of LLMs in various tasks under the Few-shot CoT setting. This reveals the relative difficulty LLMs have in understanding and responding to factual questions in different tasks, providing insights for future training of factual knowledge in LLMs. From Table \ref{domain-result}, it is observed that LLMs exhibit relatively poorer performance on factual questions related to the real-world, domain-specific knowledge, and multilingualism, being on average 6.4\% lower compared to the other four tasks. This is attributed to the fact that the training data for LLMs typically come from general domains and are not up-to-date, which indirectly inspires the exploration of retrieval-augmented LLMs \citep{ram2023context}. We analyze the LLMs in different tasks in Sec. \ref{analysis_section}.

% The benchmark proposed in this paper establishes a novel dataset and evaluation methodology to assess the factual understanding capability of LLMs. The accuracy of these models in fact-checking is measured by evaluating their performance on our dataset. 

% The model's input consists of various forms of prompts along with a claim. The output is then mapped to a label classification to determine its correctness. In the figure, ``Q'' represents a form of input provided to GPT-3.5-Turbo, and ``A'' represents the response generated by GPT-3.5-Turbo. Based on the answer provided by GPT-3.5-Turbo (A), we can map the result to ``SUPPORTS'' indicating that the model's output aligns with the label corresponding to the claim. This demonstrates that GPT-3.5-Turbo provides the correct output for this input. If the results are inconsistent with the label given by dataset, it indicates an incorrect output. Therefore, the model's accuracy on the dataset can be calculated.

% \input{tables/reflection}
% \input{tables/verification}

\begin{table}[t!]
  \caption{Results of different LLMs using Few-shot w/ CoT prompts across different tasks.}
  \label{domain-result}
  \centering
  \scalebox{0.74}{
\begin{tabular}{ccccccccccccccccc}
\toprule
\multirow{2}{*}{Task} & \multicolumn{2}{c}{Multifaceted} & \multicolumn{2}{c}{Structural} & \multicolumn{2}{c}{Adversarial} & \multicolumn{2}{c}{Temporal}  & \multicolumn{2}{c}{Real-World} & \multicolumn{2}{c}{Domain Specific} & \multicolumn{2}{c}{Multi-lingual} \\\cmidrule(lr){2-3} \cmidrule(lr){4-5} \cmidrule(lr){6-7}  \cmidrule(lr){8-9} \cmidrule(lr){10-11}\cmidrule(lr){12-13}\cmidrule(lr){14-15}
                        & Acc.            & F1             & Acc.           & F1            & Acc.           & F1             & Acc.          & F1            & Acc.           & F1            & Acc.             & F1               & Acc.            & F1              \\ \midrule
OPT-6.7B                & 34.5            & 24.1           & 45.5           & 30.9          & 51.8           & 51.7           & 30.0          & 18.0          & \textbf{53.7}  & 27.5          & 28.2             & 28.3             & 16.2            & 17.7            \\
BLOOM-7B                & 10.7            & 13.5           & 0.8            & 3.5           & 2.0            & 3.7            & 3.7           & 7.7           & 5.4            & 8.5           & 11.8             & 15.6             & 9.8             & 15.9            \\
LLaMA-7B                & 38.3            & 33.9           & 44.1           & 32.1          & 43.2           & 46.1           & 41.6          & 30.0          & 26.4           & 26.3          & 23.6             & 25.0             & 27.8            & 27.7            \\ \midrule \midrule
Alpaca-7B               & 38.6            & 28.8           & 48.0           & 23.6          & 46.4           & 35.1           & 49.6          & 26.1          & 24.5           & 19.9          & 42.9             & 26.8             & 24.2            & 17.7            \\ 
Vicuna-7B               & 44.2            & 36.0           & 49.7           & 36.3          & 59.0           & 59.2           & \textbf{50.1} & 37.6          & 49.0           & 41.8          & \textbf{44.3}    & 38.6             & \textbf{46.7}   & 43.1            \\
Vicuna-13B              & 49.9            & 45.3           & 48.1           & 37.9          & 58.9           & 60.0           & 45.4          & \textbf{37.8} & 47.7           & 42.7          & 43.5             & 40.4             & 37.8            & 37.9            \\
ChatGLM-6B              & 41.0            & 36.0           & 46.8           & 35.7          & 51.5           & 48.6           & 39.4          & 32.4          & 48.9           & 34.8          & 35.2             & 35.0             & 37.1            & 35.3            \\
Flan-T5-11B             & 49.2            & 49.4           & 43.5           & 33.7          & 54.7           & 56.6           & 31.6          & 30.6          & 31.1           & 29.4          & 35.6             & 34.6             & 25.3            & 14.4            \\ \midrule \midrule
Text-Davinci-002        & 47.7            & 47.7           & \textbf{50.8}  & \textbf{38.4} & 64.2           & 64.3           & 33.9          & 31.1          & 51.7           & 41.4          & 36.4             & 36.1             & 43.1            & 39.5            \\
Text-Davinci-003        & 51.1            & 47.8           & 44.3           & 33.7          & 64.1           & 63.7           & 41.4          & 35.1          & 48.0           & 42.8          & 40.4             & \textbf{41.4}    & 43.7            & \textbf{43.6}   \\ 
GPT-3.5-Turbo           & \textbf{53.6}   & \textbf{53.1}  & 44.8           & 37.8          & \textbf{67.4}  & \textbf{67.4}  & 37.4          & 33.9          & 50.4           & \textbf{43.1} & 38.7             & 40.3             & 41.3            & 41.1                \\
\bottomrule
\end{tabular}
 }
  \vspace{-7mm}
\end{table}
\subsection{Analysis}\label{analysis_section}
% In this section, we delve into various aspects of LLMs to explore the capabilities and limitations inherent to them. We focus on several key areas, including the manner in which LLMs manage complex reasoning tasks that involve multi-hop factual questions, the proficiency of LLMs in managing diverse data formats and storage, and their ability to tackle challenges such as numerical reasoning and entity ambiguity. Additionally, we examine how well they can analyze time-sensitive factual questions, withstand adversarial attacks, and perform when confronted with fine-grained labels and prompts in multiple languages. A detailed analysis is as follows:

In this section, we explore LLMs' capabilities focusing on key areas like handling of multi-hop factual questions, proficiency in diverse prompt strategies, and tackling challenges like numerical reasoning and entity ambiguity. We also examine their performance on time-sensitive factual questions, against adversarial attacks, with fine-grained labels and prompts in multiple languages. 

\begin{figure}[h]
  \label{fig:2}
  \begin{minipage}[]{0.33\linewidth}
    \subfigure[Multi-hop Reasoning Analysis]{
        \label{fig:2a}
        \includegraphics[width=\linewidth]{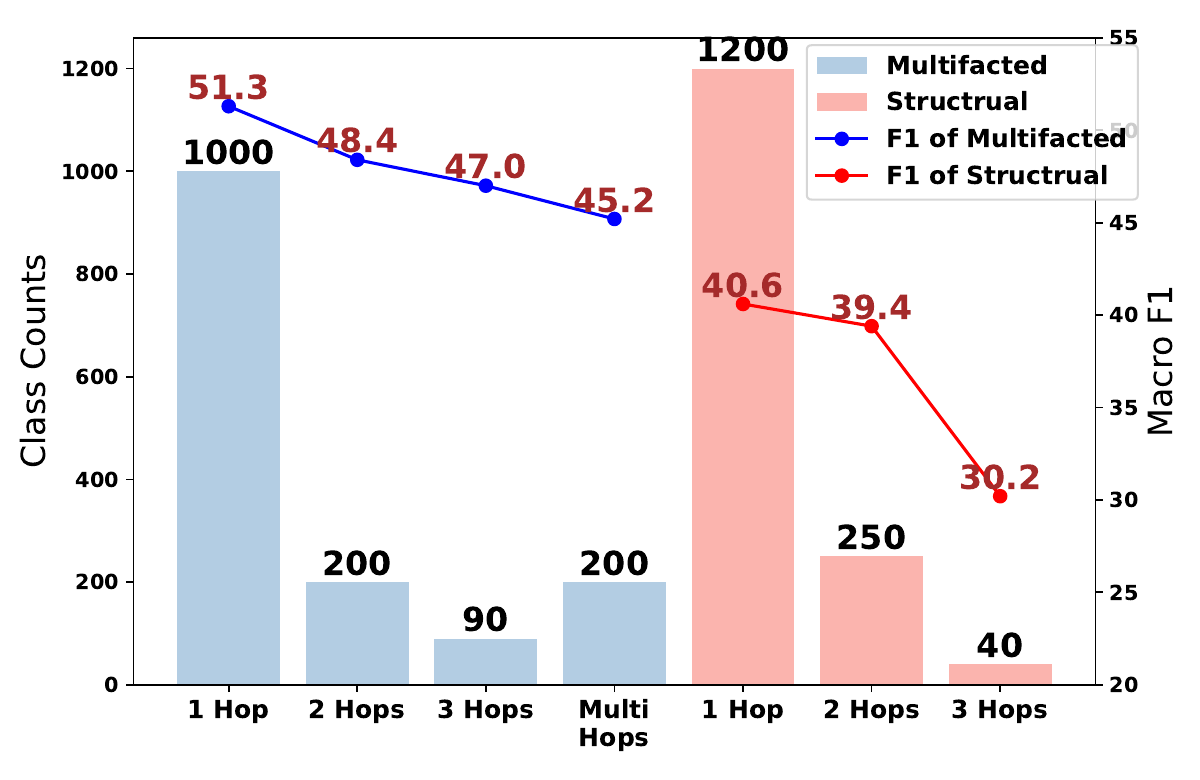}
    }
  \end{minipage}
  \begin{minipage}{0.33\linewidth}
    \subfigure[Structural Knowledge Analysis]{
        \label{fig:2b}
    
        \includegraphics[width=\linewidth]{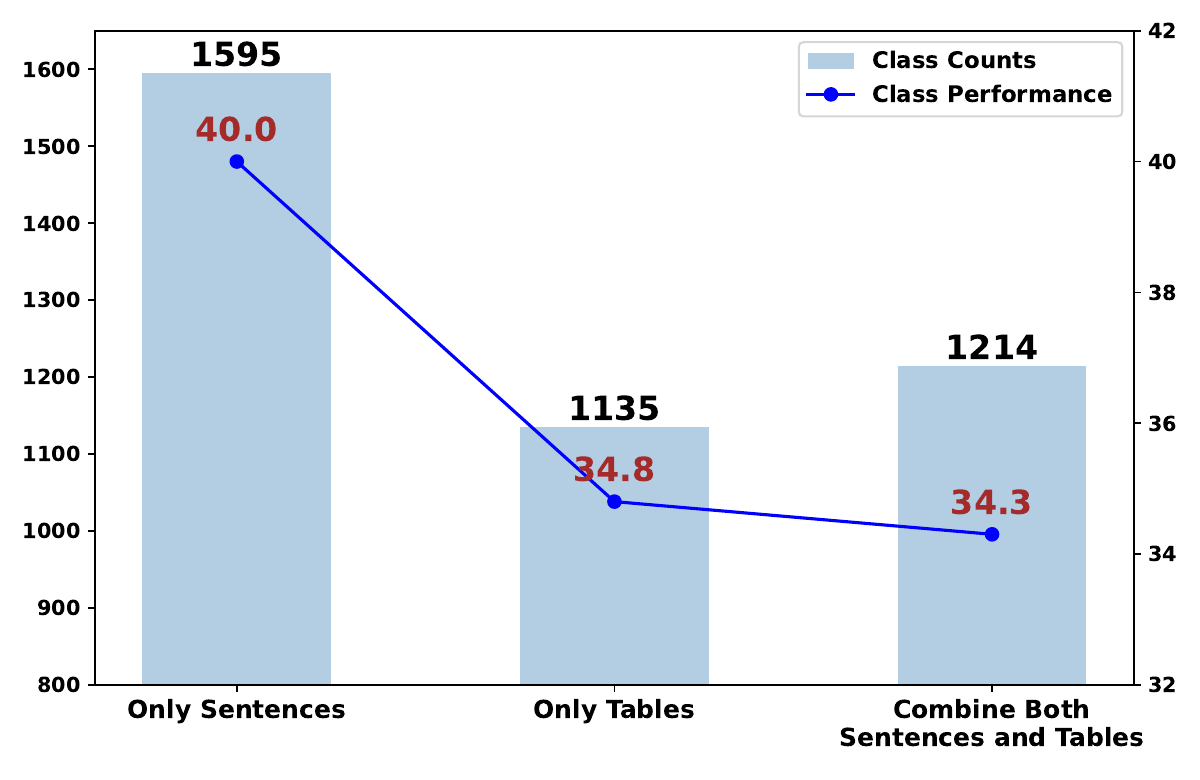}
    }  
    \end{minipage}
  \begin{minipage}{0.33\linewidth}
        \subfigure[Challenges of Different Questions]{
        \label{fig:2c}
        
        \includegraphics[width=\linewidth]{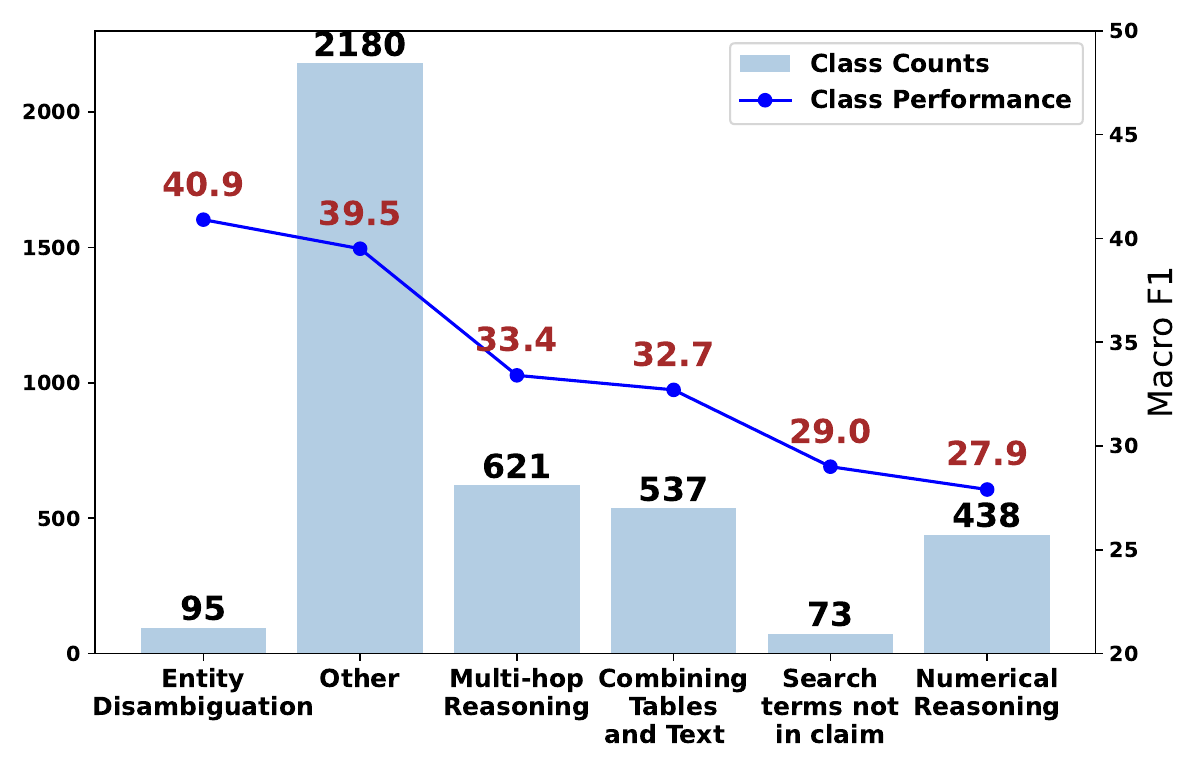}
    }  
  \end{minipage}
  \vspace{-5mm}
  \caption{GPT-3.5-Turbo's outcomes across three distinct tasks under Few-shot CoT setting.}
  \vspace{-6mm}
\end{figure}

\paragraph{Multi-hop Factual Question Analysis}
To analyze the performance of LLMs when faced with factual questions based on multiple pieces of facts that require complex logical reasoning, we categorize multifaced and structural factual questions into distinct subsets, depending on the number of ``hops'' necessary to validate each factual question. To maintain fairness, we randomly sampled 1,490 data pieces from each of the two datasets for verification. Figure ~\ref{fig:2a} illustrates the data counts and Macro F1 scores of GPT-3.5-Turbo for each respective subset. The figure reveals a clear pattern: as the number of ``hops'' increases, the reasoning chain for deriving conclusions from existing factual knowledge extends, necessitating heightened logical reasoning capabilities from the LLMs. Consequently, the performance of the LLMs exhibits diminishing trends. 

\vspace{-1mm}

\paragraph{Structural Knowledge Analysis in LLMs}

To investigate whether LLMs can effectively memorize factual knowledge from structured data, we divided the structural task questions into three subsets according to evidence distribution: evidence in unstructured data (Only text), structured data (Only tables), or both (Combine text and tables). Figure ~\ref{fig:2b} shows a notable decline (Avg. -5.5\%) in GPT-3.5-Turbo's performance when evidence involves structured data, indicating LLMs' limited ability in extracting knowledge from structured tables. The LLMs also perform less effectively when handling questions requiring the combination of both evidence types, reflecting their incapacity to integrate diverse structured evidence effectively.

\vspace{-1mm}

\paragraph{Analysis of Different Factual Questions Poses Challenges}

To assess the capabilities of LLMs in addressing various challenges, we partitioned each factual question within the structural task into six distinct challenges: 1) Entity disambiguation, 2) Other,
3) Multi-hop reasoning, 4) Combining tables and text, 5) Search terms not in claim, 6) Numerical reasoning, each centered around the most critical difficulty encountered during verification. Figure ~\ref{fig:2c} illustrates GPT-3.5-Turbo's performance and data distribution across challenges. The extensive training and large-scale parameters enhance LLMs' performance in handling entity ambiguity. Longer reasoning chains and various forms of evidence challenge LLMs’ factual abilities. When correct inference involves unmentioned entities, LLMs may lack necessary hints from factual questions, posing significant challenges. LLMs also exhibit deficiencies in precise numerical calculations due to the inherent hallucination phenomenon, resulting in subpar performance when numerical reasoning is needed for verification.

\begin{figure}[h]
  \label{fig:analysis2}
  \begin{minipage}[]{0.33\linewidth}
    \subfigure[Temporal Questions Verification]{
        \label{fig:analysis2a}
        \includegraphics[width=\linewidth]{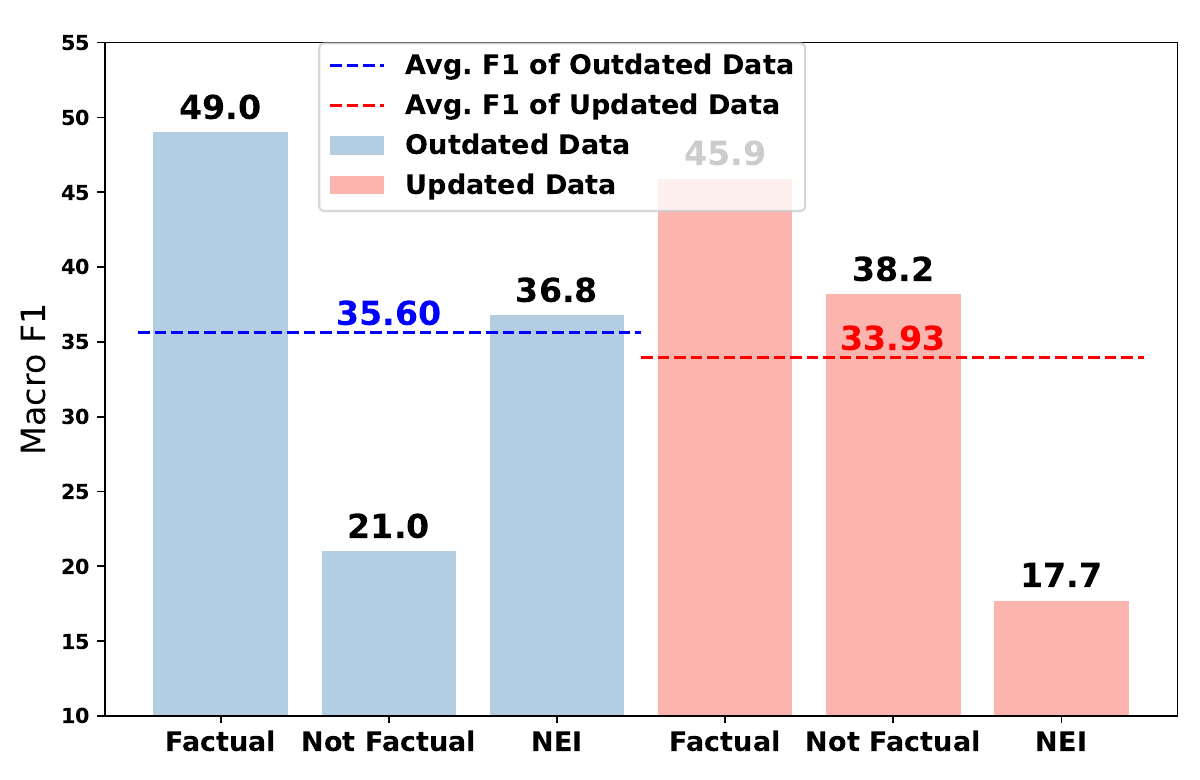}
    }
  \end{minipage}
  \begin{minipage}{0.33\linewidth}
    \subfigure[Adversarial Attacks Resilience]{
        \label{fig:analysis2b}
    
        \includegraphics[width=\linewidth]{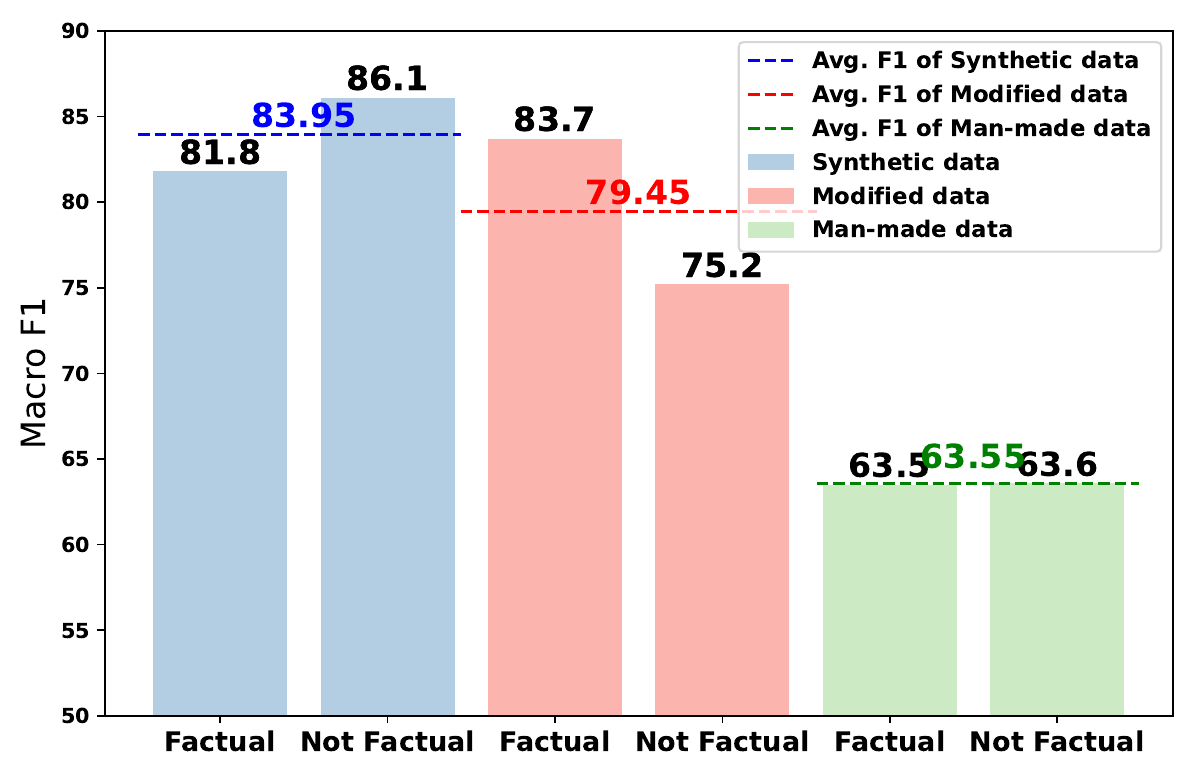}
    }  
    \end{minipage}
  \begin{minipage}{0.33\linewidth}
        \subfigure[Label Granularity Variations]{
        \label{fig:analysis2c}
        
        \includegraphics[width=\linewidth]{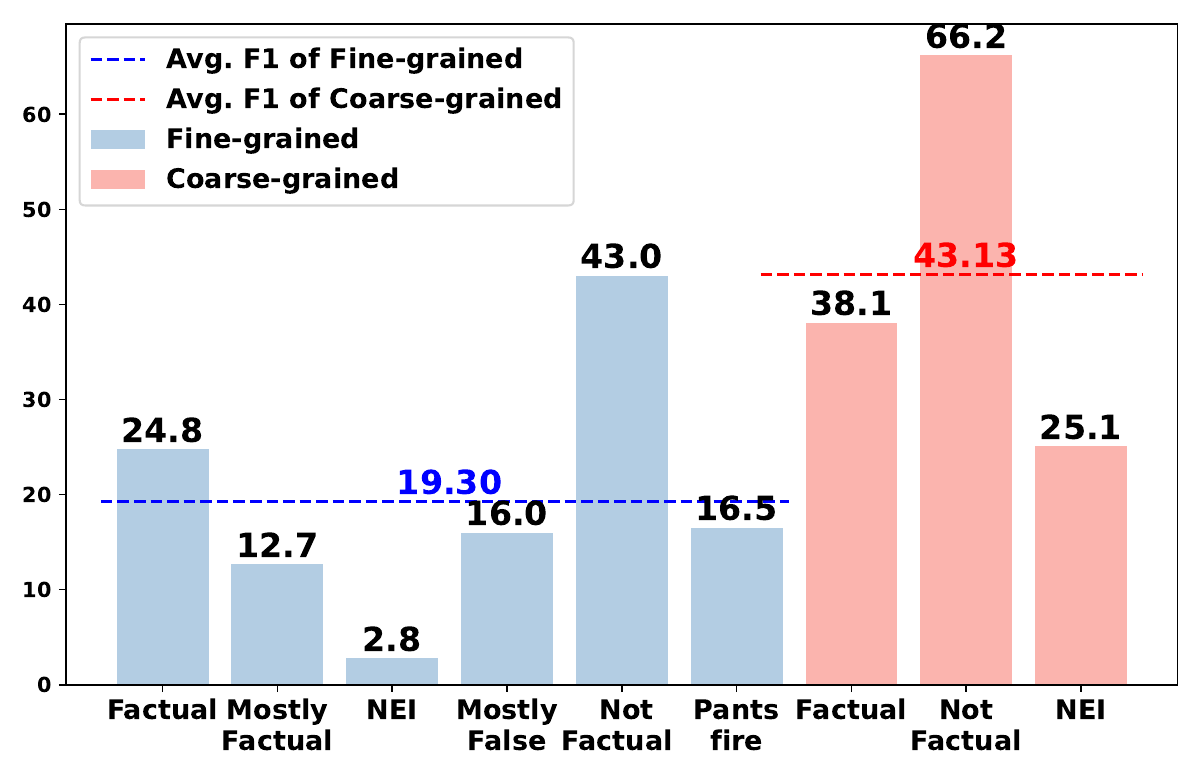}
    }  
  \end{minipage}
  \vspace{-5mm}
  \caption{Results of GPT-3.5-Turbo in three different tasks under Few-shot CoT setting.}
  \vspace{-6mm}
\end{figure}

\paragraph{Temporal Analysis}
As time progresses, the truthfulness of certain questions may undergo changes. The temporal task encompasses such data, and we leverage this task to explore the ability of LLMs to adapt to factual changes over time. Figure ~\ref{fig:analysis2a} illustrates that GPT-3.5-Turbo exhibits superior performance when dealing with outdated data as compared to updated data. This discrepancy arises from the fact that LLMs are pretrained on a corpus of text prior to a specific temporal point. Consequently, LLMs lack the capability to acquire real-time, up-to-date knowledge, rendering them unable to validate questions that hinge on the most recent information for accurate assessments.

\paragraph{Adversarial Analysis}
To evaluate the robustness of LLMs to adversarial attacks, we divide the adversarial questions into three subsets: auto-generated questions from the corpus, manually modified synthesized questions yielding adversarial ones, and artificially created adversarial questions.

Figure ~\ref{fig:2b} presents the performance of GPT-3.5-Turbo on these three subsets. It is evident that following adversarial attacks, LLMs exhibit a substantial decrease in performance. Furthermore, factual questions that have undergone manual modifications or were artificially created prove to be more challenging compared to those that are automatically generated ~\citep{Shen2023}. This disparity could be attributed to the fact that automatically synthesized factual questions often contain explicit positive or negative words that hint at the outcome, and the exceptional comprehension abilities of LLMs enable them to accurately discern and provide the correct response in such cases.

\paragraph{Label Granularity Analysis}
To assess the effect of different label granularities on LLMs' performance, we conducted a manual re-labeling of the real-world task questions. Per the settings of \citet{Poltifact2022}, besides labeling as ``Factual'', ``Non-Factual'', and ``Not Enough Information'', we also require them to annotate the dataset with six factual labels: ``Factual'',  ``Mostly Factual'', ``Mostly False'', ``Non-Factual'', ``Pants-Fire'', and ``Not Enough Information''. We also modified the prompt for GPT-3.5-Turbo for more intricate factual responses to test its competency with nuanced labels. Results in Figure \ref{fig:analysis2c} disclosed: 1) The results show that, in general, there is a significant decrease in performance (-23.83\%) when transitioning from coarse-grained justification to fine-grained justification. With finer granularity, LLMs are not only required to assess the authenticity of each question but also to judiciously employ their knowledge base to precisely gauge the credibility of each factual questions. 2) When comparing the performance of coarse-grained labels with fine-grained labels, we observe significant drops in the three categories: ``Factual'' by 13.3\%, ``Non-Factual'' by 23.2\%, and ``Not Enough Information'' by 22.3\%. This indicates that finer-grained labels introduce additional options that can potentially disrupt the original judgment of the LLMs. A potential remedy could be the aggregation of multiple judgments through voting ~\citep{wang2023selfconsistency}.

\paragraph{Multilingual Task with Chinese and English Prompts}
\begin{wraptable}
{r}{0.3\linewidth}
    \scalebox{0.77}{
\begin{tabular}{ccc}
\toprule
Language    & English  & Chinese  \\
\midrule
Factual     & 41.7           & \textbf{55.5}           \\
Non-Factual & 47.9           & \textbf{49.7}           \\
NEI         & \textbf{43.8}           & 35.5           \\
Overall     & 44.5           & \textbf{46.9}           \\
\bottomrule
\end{tabular}}

    \caption{Macro F1 over Chinese and English prompts.}
    \label{table:language}
    \vspace{-6mm}
\end{wraptable} 
To investigate the influence of prompts in different languages on LLMs, we extracted Chinese factual questions from the multilingual tasks to create a subset. We then evaluated the LLMs' performance when using both Chinese and English prompts, both of which are depicted in Appendix  ~\ref{Prompt_format}. Table ~\ref{table:language} illustrates the results,
indicating that the LLMs perform better when using a Chinese prompt. This underscores the notion that employing prompts in the same language as the questions can enhance the transfer capabilities from English factual knowledge to other languages of LLMs.

% \newline
\vspace{-5mm}
\begin{table}[h]
  \caption{Results in different domains obtained on the Pinocchio-Lite using different prompts.}
  \label{lite-result}
  \centering
  \scalebox{0.67}{
\begin{tabular}{ccccccccccccccccc}
\toprule
\multirow{2}{*}{Task} & \multicolumn{2}{c}{Multifacted} & \multicolumn{2}{c}{Structural} & \multicolumn{2}{c}{Adversarial} & \multicolumn{2}{c}{Temporal} & \multicolumn{2}{c}{Real-World} & \multicolumn{2}{c}{Domain Specific} & \multicolumn{2}{c}{Multi-lingual} & \multicolumn{2}{c}{Overall} \\
\cmidrule(lr){2-3} \cmidrule(lr){4-5} \cmidrule(lr){6-7} \cmidrule(lr){8-9} \cmidrule(lr){10-11} \cmidrule(lr){12-13} \cmidrule(lr){14-15} \cmidrule(lr){16-17}

                        & Acc.           & F1             & Acc.           & F1            & Acc.           & F1             & Acc.          & F1           & Acc.           & F1            & Acc.             & F1               & Acc.            & F1              & Acc.         & F1           \\ \midrule

1 shot                  & 56.0           & 50.9           & 37.0           & 35.7          & 50.5           & 56.6           & 39.5          & 39.5         & 43.0           & 42.7          & 40.0             & 40.1             & 42.0            & 38.7            & 44.0         & 43.7         \\
2 shots                 & 56.0           & \textbf{53.4}           & 41.0           & 42.3          & 47.5           & 56.2           & 41.0          & 42.0         & 40.5           & 41.7          & \textbf{42.5}             & \textbf{43.5}             & 36.5            & 34.8            & 43.6         & 43.7         \\
3 shot (Ours)                & 54.5           & 50.0           & 38.0           & 36.8          & 49.0           & 54.9           & 40.0          & 39.0         & 39.5           & 38.1          & 41.5             & 41.7             & 40.5            & 39.2            & 43.3         & 43.9         \\
6 shots                 & 54.5           & 51.7           & 38.5           & 38.3          & 49.0           & 55.8           & 42.0          & 41.5         & 42.5           & 41.6          & 39.0             & 39.5             & 41.0            & 38.4            & 43.8         & 43.8         \\
9 shots                 & \textbf{57.5}           & 53.3           & 38.0           & 37.8          & 52.0           & 57.3           & 43.0          & 42.2         & 42.5           & 39.8          & 37.5             & 36.7             & 37.5            & 35.0            & 44.0         & 44.0         \\
12 shots                & 55.5           & 52.0           & 38.5           & 38.6          & 53.0           & 58.8           & \textbf{47.0}          & \textbf{46.9}         & \textbf{46.0}           & \textbf{44.7}          & 34.0             & 34.5             & 39.0            & 37.1            & 44.7         & 44.8         \\ 
\midrule
\midrule

Complex chain             & 51.0           & 50.2           & 38.5           & 35.0          & 37.5           & 47.2           & 39.0          & 39.0         & 39.5           & 36.8          & 36.0             & 35.7             & 38.5            & 31.7            & 40.0         & 39.7         \\ 
Self-consistency             & 55.5           & 51.2           & 43.0           & 42.6          & 49.5           & 54.8           & 43.0          & 41.6         & 43.0           & 41.9          & 42.0             & 42.4             & 39.5            & 36.8            & 45.1         & 45.0    
\\
Self-refinement        &                55.0&                52.1&                \textbf{44.5}&               \textbf{44.0}&                \textbf{53.5}&                \textbf{59.2}&               42.5&              42.2&                41.5&               40.3&                  42.0&                  43.4&                 \textbf{43.0}&                 \textbf{39.9}&              \textbf{46.0}&              \textbf{46.2}\\
Declarative Claim        &                52.0&                51.1&                39.0 &      35.1         &      45.5          &    49.3            &               40.5&              40.7&                40.0&               37.9&                  41.0&                  40.6&                 38.5&                 36.3&       42.3       &        41.6      \\
\bottomrule
\end{tabular}
  }
  \vspace{-5mm}
\end{table}

\paragraph{Prompt Strategy Analysis}\label{prompt_strategy_analysis}

In prior research, various CoT methods have been employed to enhance the performance of LLMs. These methods include 1) augmenting the number of in-context learning examples, 2) implementing self-consistency mechanisms, which alleviates the hallucination phenomenon through majority voting after multiple judgments of LLMs ~\citep{wang2023selfconsistency}, 3) incorporating complex reasoning chains, which leverages the most complex CoT in prompt to steer the cognitive processes of LLMs and augment their cognitive capabilities ~\citep{fu2022complexity}, and 4) employing self-refinement strategies, which refines LLMs' answers through continuous feedback of another LLM on responses to achieve better results ~\citep{madaan2023self} and so forth. Additionally, we examined the influence of utilizing declarative claims as instances of in-context learning. We randomly sampled 200 factual questions from each task of the Pinocchio, totaling 1400 questions, to compose Pinocchio-Lite with the aim of speeding up the testing of different prompt strategies. The performance results of various CoT methods are presented in Table \ref{lite-result}. To maintain fairness, three in-context learning examples are employed in the complex chain, self-consistency, self-refinement, and declarative claim methods. Different types of CoT prompts are shown in Appendix \ref{Prompt_format}.

It is worth noting that 1) when the number of in-context learning examples is limited, the incremental improvement in performance is marginal upon increasing the number of examples. However, beyond a specific threshold, the addition of more examples gains more performance improvement. This could be due to the inability of LLMs to fully encapsulate the correct reasoning with fewer examples. 2) Concurrently, a fascinating observation is that the LLM’s performance substantially deteriorates as the complexity of the CoT increases. This could stem from the difficulty LLMs have in extracting a generalized reasoning pattern from complex, multi-stage thinking processes with limited examples. 3) The self-consistency method markedly boosts performance by mitigating the hallucination issue in LLMs through consistency voting, enhancing their response accuracy. 4) In the self-refinement approach, the model might initially provide an incorrect response, but it can amend its mistakes through feedback and refine its answers. In the end, when no additional refinement is needed, the model often reaches the correct conclusion, achieving optimal performance. 5) Compared to the 3 shots method, the declarative claims method saw a 2.3\% performance drop, illustrating that using questions as examples effectively directs LLMs in acquiring factual knowledge.

\section{Related Work}

\paragraph{Factual Knowledge in Language Models} 

% An important question is whether pretrained MLMs know facts about real-world entities. The LAMA dataset~\citep{PetroniRRLBWM19} evaluates this using cloze tests that consist of (sub, rel, obj) triples, e.g. (Obama, bornIn, Hawaii), and manually created prompts with missing objects, e.g. “Obama was born in [MASK].”. LPAQA~\citep{JiangXAN20} extends this idea by systematically creating prompts that are generated by mining Wikipedia, paraphrasing, and crowdsourcing.

Previous research has demonstrated that LLMs have the ability to retain and utilize factual knowledge, effectively acting as knowledge bases~\citep{PetroniRRLBWM19, PetroniLPRWM020, HeinzerlingI21}. This acquired factual knowledge in language models during pretraining can be advantageous for knowledge-intensive tasks like question answering and fact checking~\citep{RobertsRS20,wenhao2023,PanWLLWKN23}. To evaluate the factual knowledge stored in language models, \citet{PetroniRRLBWM19} employed cloze tests consisting of triples and prompts specifically designed to simulate missing objects. \citet{JiangAADN20} explored the role of prompts in retrieving factual information from language models and devised improved prompts for probing. However, \citet{ElazarKRRHSG21} demonstrated the unreliability of rank-based probing methods with paraphrased context, leading to inconsistent findings. \citet{CaoLHSYLXX20} contended that biased prompts and leakage of golden answers often lead to overestimations of LLMs' knowledge storage capability. In contrast, \citet{VarshneyMB22a} used question answering to measure models' uncertainty regarding specific facts. Our method is more in line with ~\citet{Kadavath2022} and \citet{LinHE22}, employing self-evaluation by querying the models to assess response accuracy regarding factual knowledge.

\paragraph{Benchmarks for Large Language Models} 
The advent of LLMs has underscored the importance of exhaustive benchmarks for effective capability assessment. Presently, there are predominantly two types of existing benchmarks. One evaluates the general knowledge and reasoning capacities of LLMs, exemplified by the MMLU benchmark~\citep{HendrycksBBZMSS21}, a multi-task evaluative measure encompassing tasks from real-world tests and literature, spanning diverse subjects like elementary math, US history, computer science, and law. Moreover, benchmarks also exist for non-English languages~\citep{Huang2023} or in a bilingual context~\citep{Zhong2023}. BIG-bench~\citep{Srivastava2022} is a collaborative benchmark examining LLMs' capabilities across 204 diverse tasks from various fields like linguistics, childhood development, software development, and more. HELM~\citep{Liang2022} employs 7 metrics over 42 tasks to assess LLMs, focusing on aspects from accuracy to robustness. Specific benchmarks like GSM8K~\citep{Cobbe2021} and MATH~\citep{HendrycksBBZMSS21} target mathematical problem-solving, presenting elementary to competition-level problems. In program synthesis, HumanEval~\citep{chen2021} and MBPP~\citep{Austion2021} evaluate functional correctness through program synthesis from docstrings. Additional benchmarks address instruction following~\citep{Dubois2023}, tool usage~\citep{xu2023tool}, and decision making~\citep{liu2023agentbench}. Our benchmark mainly evaluates factual knowledge, differing from ones like TruthfulQA~\citep{LinHE22}, which specifically tests truthfulness in LLMs' generated responses, with questions structured to provoke imitative falsehoods over truthful answers.

\section{Conclusion}

In this work, we investigate whether LLMs are capable of memorizing factual knowledge and reasoning based on it, across various problem categories and prompting strategies.
To this end, we curate the Pinocchio benchmark, a comprehensive test bed with 20,713 questions covering seven tasks with varying complexity. By evaluating
LLMs and prompting approaches on the Pinocchio benchmark, we find that different types of LLMs employing various prompting strategies, such as multi-shots and self-consistency, still perform suboptimally on factual tasks. Improving LLMs’ factual knowledge and reasoning abilities on complex and nuanced NLP tasks remains an open research question, and we encourage future work to develop upon our proposed Pinocchio benchmark.

\bibliography{iclr2024_conference}
\bibliographystyle{iclr2024_conference}

\appendix
\section{Appendix}

\subsection{Ethical Statement}

Pinocchio primarily serves to assess LLMs' responses to questions concerning factual knowledge. If a model performs effectively, it would be imprudent to infer that its reliability will uniformly translate to diverse task domains (even if some degree of transfer learning is anticipated). For instance, Pinocchio does not encompass long-form generation, such as news articles, or interactive settings, such as extended dialogues with adversarial entities. Furthermore, although the questions within Pinocchio parallel real-world inquiries, they originate not from a deployed system, thus posing a potential risk of over- or under-estimating the factuality of such a system.

We postulate that Pinocchio is unlikely to prove advantageous for those intending to fabricate deceptive models with malicious intent. To effectuate deception, a model must generate erroneous responses relatively infrequently, lest humans swiftly discern its unreliability. However, acquiring a low score on Pinocchio necessitates the provision of incorrect answers to virtually all questions. To be instrumental for malevolent purposes, a model must generate highly specific false statements, such as assertions concerning a maliciously targeted victim or a particular governmental policy. Yet, Pinocchio lacks coverage of highly specific subjects, offering instead a superficial overview of general factual topics.

While Wikipedia and some news websites are exemplary collaborative resources, they inherently contain inaccuracies and noise, akin to any encyclopedia or knowledge repository. Consequently, we advise users of Pinocchio against making absolute assertions about the validated claims and discourage its utilization for the development of truth-revealing models. We refrained from collecting participants' personal data in any form. Participants accessed our online tool exclusively using an identification number. Generated assertions must solely incorporate information deemed as general world knowledge or sourced from Wikipedia, thereby excluding any personally identifiable information or offensive content.

\subsection{The detailed introduction to the LLMs}\label{LLM_INTRO}

For pretraining models, OPT~\citep{OPT2022} is an open-sourced large causal language model which perform similar in performance to GPT-3~\citep{BrownMRSKDNSSAA20}. BLOOM~\citep{Bloom2022} is an open-access multilingual large language model that is suitable for non-English facts. LLaMA~\citep{touvron2023} is probably the best open-weight foundation model so far that achieves the highest accuracy on various English benchmarks (e.g. MMLU~\citep{HendrycksBBZMSS21}) within open-weight models. For instruction-tuned models, Alpaca~\citep{alpaca2023} is fine-tuned from the LLaMA model on 52K self-instructed demonstrations~\citep{WangKMLSKH23}. Alpaca behaves qualitatively similarly to OpenAI’s Text-Davinci-003 on evaluation of single-turn instruction following. Vicuna is an open-source chatbot trained by fine-tuning LLaMA on user-shared conversations collected from ShareGPT~\citep{sharegpt2023}. Flan
-T5~\citep{flan2022} is an enhanced version of T5 that has been instruction fine-tuned in a mixture of tasks. ChatGLM is an open bilingual language model based on the General Language Model~\citep{ZengLDWL0YXZXTM23}. ChatGLM is trained on Chinese and English corpus, supplemented by instruction tuning, feedback bootstrap, and reinforcement learning with human feedback (RLHF; \citealt{Ouyang0JAWMZASR22}). ChatGPT~\citep{chatgpt2022} from OpenAI that has undergone pretraining, instruction tuning, and RLHF.  ChatGPT has been observed to have impressive capabilities in various aspects favoring reasoning capabilities~\citep{Qin2023}.

\subsection{Task Results}
In this section, we present the results of all LLMs across different tasks under three different settings: Zero-shot w/o CoT, Zero-shot w/ CoT, and Few-shot w/o CoT.
\newpage
\begin{table}[htbp!]
  \caption{Results of different LLMs using Zero-shot w/o CoT prompts across different domains.}
  \label{domain-result-zeroshot-no-cot}
  \centering
  \scalebox{0.74}{
\begin{tabular}{ccccccccccccccccc}
\toprule
\multirow{2}{*}{Task} & \multicolumn{2}{c}{Multifaceted} & \multicolumn{2}{c}{Structural} & \multicolumn{2}{c}{Adversarial} & \multicolumn{2}{c}{Temporal}  & \multicolumn{2}{c}{Real-World} & \multicolumn{2}{c}{Domain Specific} & \multicolumn{2}{c}{Multi-lingual} \\ \cmidrule(lr){2-3} \cmidrule(lr){4-5} \cmidrule(lr){6-7}  \cmidrule(lr){8-9} \cmidrule(lr){10-11}\cmidrule(lr){12-13}\cmidrule(lr){14-15}
                        & Acc.            & F1             & Acc.           & F1            & Acc.           & F1             & Acc.          & F1            & Acc.           & F1            & Acc.             & F1               & Acc.            & F1              \\ \midrule
OPT-6.7B               & -           & -           & -           & -          & -           & -           & -          & -          & -  & -          & -             & -         & -           & -            \\  
BLOOM-7B                & 21.9            & 17.8           & 24.9            & 17.9           & 32.4            & 36.3            & 17.6           & 14.2           & 52.1            & 23.8           & 30.1              & 29.9             & 29.0             & 30.4            \\
LLaMA-7B                & 30.7            & 28.8           & 38.3           & 29.3          & 30.8           & 35.6           & 37.9          & 26.0          & 35.1          & 32.4          & 27.1             & 29.1             & 13.9            & 17.2            \\ \midrule \midrule
Alpaca-7B               &      34.8       &     21.6      &      47.9      &      23.7     &    47.7       &      35.7     &     \textbf{52.9}     &      26.8     &      28.1      &     19.0     &    \textbf{43.1}          &    24.2         &      26.4      &     19.5        \\ 
Vicuna-7B               & 38.6          & 35.4           & 19.4           & 16.8          & 50.8           & 53.9           & 37.9 & \textbf{42.0}          & 29.8           & 30.1          & 33.6   & 30.4             & 34.8   & 34.4            \\
Vicuna-13B              & 45.0            & 41.1           & 43.9           & 31.0         & 57.1           & 56.7           & 45.9          & 33.7  &  32.0           & 29.0          & \textbf{43.1}             & 32.3             & 37.3           & 34.7                \\
ChatGLM-6B              & 30.6            & 30.3           & 45.6           & 30.8          & 42.9           & 46.4          & 28.0          & 24.1         & 45.9           & 31.9          & 34.1             & 30.2             & 32.9            & 28.5            \\
Flan-T5-11B             & 39.2            & 29.6           & 11.2           & 10.2          & 56.2           & 49.9           & 12.9          & 10.5          & 17.4           & 10.6          & 28.8             & 16.5             & 25.4            & 14.7            \\ \midrule \midrule
Text-Davinci-002        &     44.7        &    38.4     &        \textbf{49.2}    &      \textbf{37.8}     &     57.2      &      56.1     &      36.2     &      27.8     &      \textbf{53.2}      &     32.7     &         31.3     &     30.1        &    42.2        &     32.5           \\
Text-Davinci-003         &    50.9         &     48.9      &        36.4    &     29.5      &     58.7      &     57.9      &     51.7      &      36.6     &      40.4      &    37.0      &        41.3      &      33.3       &     42.7       &  43.1      \\ 
GPT-3.5-Turbo             &      \textbf{53.2}       &      \textbf{50.1}     &       43.1     &       35.8         &     \textbf{62.3}         &   \textbf{61.8}       &     43.4      &      35.9     &     46.1       &      \textbf{42.1}    &     42.5         &     \textbf{35.6}        &   \textbf{45.0}         &       \textbf{45.7}            \\
\bottomrule
\end{tabular}
 }
\end{table}
\begin{table}[htbp!]
  \caption{Results of different LLMs using Zero-shot w/ CoT prompts across different domains.}
  \label{domain-result-zeroshot-cot}
  \centering
  \scalebox{0.74}{
\begin{tabular}{ccccccccccccccccc}
\toprule
\multirow{2}{*}{Task} & \multicolumn{2}{c}{Multifaceted} & \multicolumn{2}{c}{Structural} & \multicolumn{2}{c}{Adversarial} & \multicolumn{2}{c}{Temporal}  & \multicolumn{2}{c}{Real-World} & \multicolumn{2}{c}{Domain Specific} & \multicolumn{2}{c}{Multi-lingual} \\ \cmidrule(lr){2-3} \cmidrule(lr){4-5} \cmidrule(lr){6-7}  \cmidrule(lr){8-9} \cmidrule(lr){10-11}\cmidrule(lr){12-13}\cmidrule(lr){14-15}
                        & Acc.            & F1             & Acc.           & F1            & Acc.           & F1             & Acc.          & F1            & Acc.           & F1            & Acc.             & F1               & Acc.            & F1              \\ \midrule
OPT-6.7B                & -           & -           & -           & -          & -           & -           & -          & -          & -  & -          & -             & -         & -           & -            \\
BLOOM-7B                & 17.0            & 20.2           & 10.1            & 12.6           & 12.0            & 19.2            & 6.9           & 9.4           & 15.5            & 16.5           & 27.3              & 23.4             & 17.9             & 19.3            \\
LLaMA-7B                & 20.3            & 23.5           & 29.5           & 26.4          & 18.3           & 26.2           & 25.7          & 26.3          & 22.9          & 24.9          & 20.0             & 23.0             & 12.2            & 16.9            \\ \midrule \midrule
Alpaca-7B               &      38.3       &      28.9     &      42.7      &       22.4    &     38.6      &    36.1       &       38.0    &       23.0    &       29.7     &    23.1      &   28.5           &      21.7       &      13.5      &      15.2       \\ 
Vicuna-7B               & 29.4          & 35.8           & 45.7           & 31.6          & 4.4           & 8.3           & \textbf{49.0} & 36.6          & 15.1           & 19.6          & \textbf{47.4}   & \textbf{39.6}             & 37.9   & 33.9            \\
Vicuna-13B              & 46.7            & 42.8           & 46.2           & 32.7         & 58.8           & 58.6           & 47.3          & 34.6  &  34.1           & 31.1          & 43.6             & 33.6             & 36.0           & 33.2                \\
ChatGLM-6B              & 34.0            & 33.0           & 40.5           & 29.8          & 46.3           & 46.6          & 27.3          & 24.7          & 44.9           & 30.7          & 32.2             & 30.1             & 30.2            & 30.4            \\
Flan-T5-11B             & 49.6            & 49.1           & 19.2           & 16.8          & 58.2           & 58.2           & 21.7          & 21.8          & 20.4           & 17.1          & 30.3             & 20.8             & 25.8            & 15.6            \\ \midrule \midrule
Text-Davinci-002        &      47.2       &    40.1       &      \textbf{51.7}     &     \textbf{38.0}      &    59.9       &   58.2        &   37.2        &    30.8       &      \textbf{52.7}      &   34.4       &  29.9            &    30.3         &     42.5       &      36.6          \\
Text-Davinci-003         &   52.7          &     51.1      &      37.5      &    31.3       &    \textbf{61.0}       &      59.5     &      40.8     &      36.7     &     38.8       &     36.2     &       41.4       &      33.0       &     42.2       &   42.4     \\ 
GPT-3.5-Turbo             &      \textbf{53.3}       &     \textbf{52.1}      &    43.1        &    35.5       &    59.8       &    \textbf{61.6}       &    42.2       &    \textbf{37.7}       &   44.8         &    \textbf{43.3}      &    41.4          &    36.0         &    \textbf{43.4}        &     \textbf{45.3}              \\
\bottomrule
\end{tabular}
 }
\end{table}
\begin{table}[htbp!]
  \caption{Results of different LLMs using Few-shot w/o CoT prompts across different domains.}
  \label{domain-result-fewshot-not-cot}
  \centering
  \scalebox{0.74}{
\begin{tabular}{ccccccccccccccccc}
\toprule
\multirow{2}{*}{Task} & \multicolumn{2}{c}{Multifaceted} & \multicolumn{2}{c}{Structural} & \multicolumn{2}{c}{Adversarial} & \multicolumn{2}{c}{Temporal}  & \multicolumn{2}{c}{Real-World} & \multicolumn{2}{c}{Domain Specific} & \multicolumn{2}{c}{Multi-lingual} \\ \cmidrule(lr){2-3} \cmidrule(lr){4-5} \cmidrule(lr){6-7}  \cmidrule(lr){8-9} \cmidrule(lr){10-11}\cmidrule(lr){12-13}\cmidrule(lr){14-15}
                        & Acc.            & F1             & Acc.           & F1            & Acc.           & F1             & Acc.          & F1            & Acc.           & F1            & Acc.             & F1               & Acc.            & F1              \\ \midrule
OPT-6.7B                & 38.1            & 30.1           & 45.9           & 27.1          & 46.8           & 32.4           & 28.7          & 20.0          & 51.1  & 25.5          & 37.0             & 29.6             & -            & -           \\
BLOOM-7B                & 32.7            & 22.5           & 8.8            & 9.0           & 43.5            & 32.6            & 23.8           & 21.1           & \textbf{53.3}            & 31.4           & 29.3              & 28.4             & 22.3             & 19.3            \\
LLaMA-7B                & 34.8            & 21.9           & 40.5           & 27.0          & 47.4           & 38.4           & 45.5          & 26.9          & 22.4           & 22.0          & 39.3             & 34.3             & 32.6            & 27.0            \\ \midrule \midrule
Alpaca-7B               &      34.9       &     25.4      &    48.0        &     22.6      &    43.4       &      32.5     &     48.0      &      25.8     &      24.0      &     19.4     &  42.6            &      27.0       &      21.8      &      17.4       \\ 
Vicuna-7B               & 34.5          & 27.6           & 40.1           & 25.4          & 54.5           & 53.3           & 30.1 & 26.6          & 36.1           & 34.0          & 33.9   & 27.7             & 22.8   & 20.5            \\
Vicuna-13B              & 47.9            & 42.5           & 48.9           & 31.4         & 54.7           & 53.1           & \textbf{53.4}          & \textbf{38.6}  &  39.7           & 35.2          & \textbf{47.4}             & 34.9             & 37.7           & 36.8                \\
ChatGLM-6B              & 37.9            & 32.9           & 44.6           & 35.4          & 52.2           & 46.8           & 44.9          & 35.4          & 38.0           & 33.9          & 41.6             & \textbf{38.0}             & 34.5            & 33.8            \\
Flan-T5-11B             & 42.3            & 35.0           & 12.4           & 11.7          & 57.7           & 53.6           & 15.1          & 13.0          & 17.7           & 11.4          & 29.7             & 19.4             & 24.9            & 13.6            \\ \midrule \midrule
Text-Davinci-002      &45.4 &41.2   &    \textbf{51.4}         &     \textbf{38.4}      &      61.7      &       \textbf{61.8}    &     37.0      &     31.3      &     52.0      &     38.6      &       33.0     &     32.6     &    42.5          &      40.0                  \\
Text-Davinci-003    & \textbf{59.6}     &    43.4       &      48.1     &    33.7    &    \textbf{62.0}     &    \textbf{61.8}   &     46.4   &   36.3  &  50.6    &     43.0      &     41.7       &    36.3      &  \textbf{44.2}            &   \textbf{44.4}                              \\ 
GPT-3.5-Turbo            &       52.1      &    \textbf{48.4}       &      42.5      &     35.4      &     61.2      &     61.1      &     43.7      &     36.2      &       48.9     &      \textbf{43.2}    &     42.0         &    35.6         &    42.8        &   43.0              \\
\bottomrule
\end{tabular}
 }
\end{table}

\subsection{Prompt Strategy}\label{Prompt_format}
In this section, we provide the comprehensive versions of all the prompts utilized in both the main experiments and the subsequent analysis. We engaged native Chinese annotators to rephrase the English prompts while maintaining their semantic integrity, thus yielding Chinese prompts. 

\newpage
\begin{figure}[htbp!]\label{Complete_prompts}
  \centering
  \includegraphics[scale=0.62]{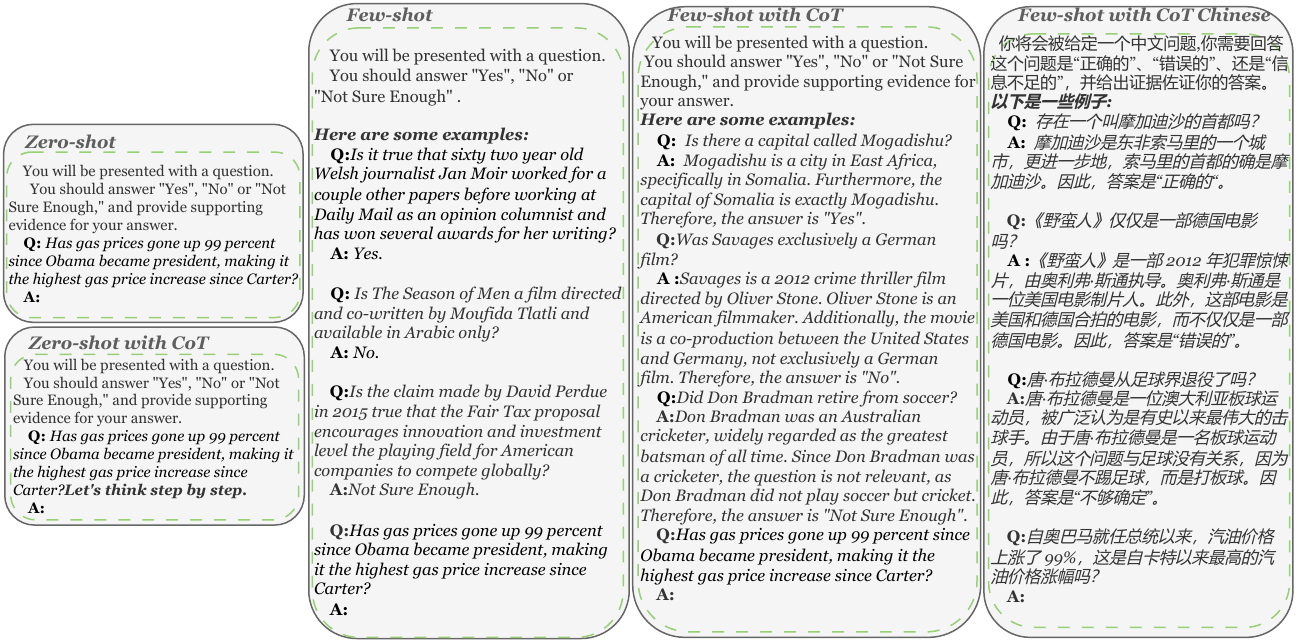}
  \caption{Prompts of four different settings.}
\end{figure}

\begin{figure}[htbp!]
  \centering
  \includegraphics[scale=0.47]{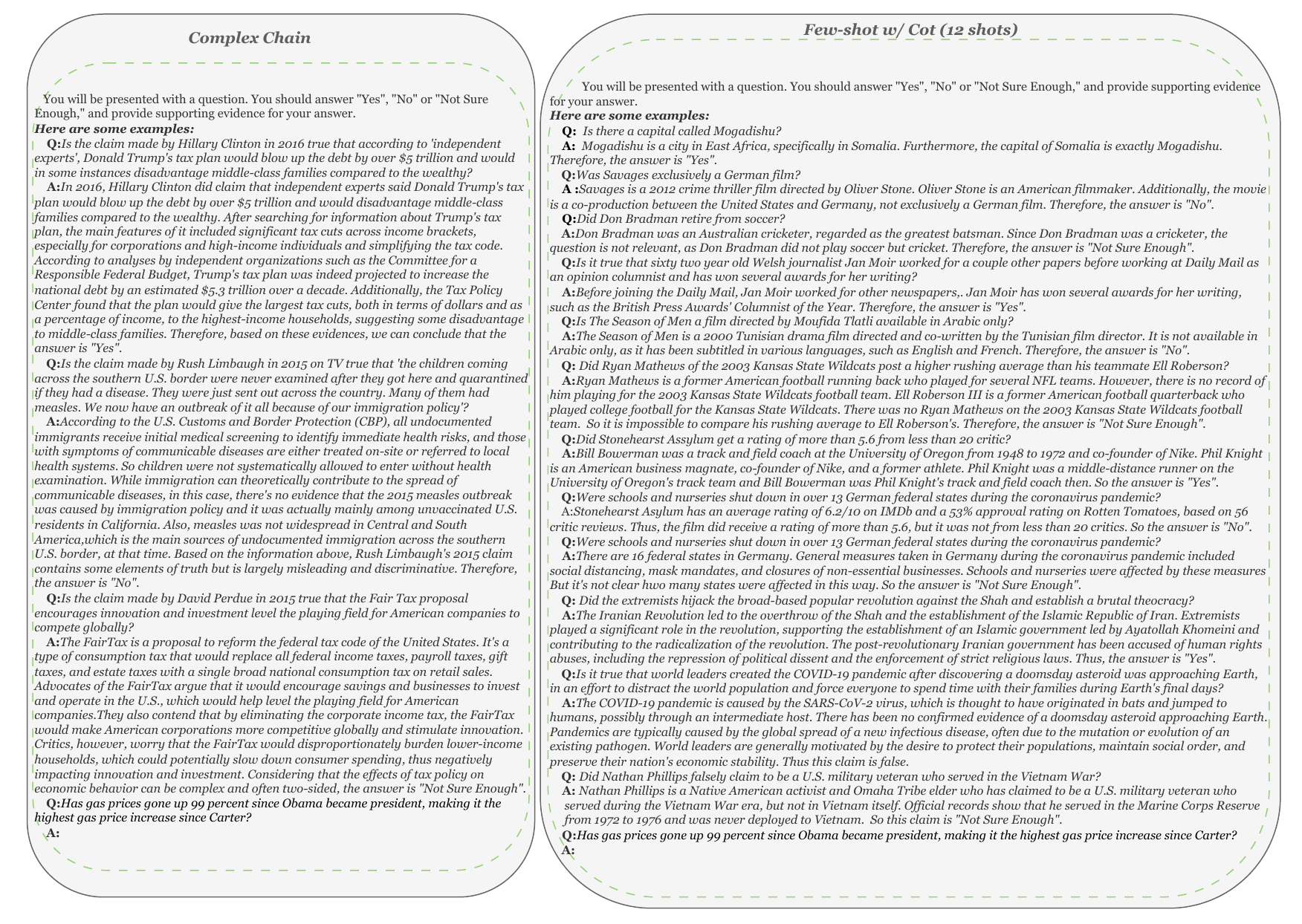}
  \caption{Prompts of complex chain and Few-shot CoT with 12 shots method.}
\end{figure}

\begin{figure}[htbp!]
   \centering
   \includegraphics[scale=0.62]{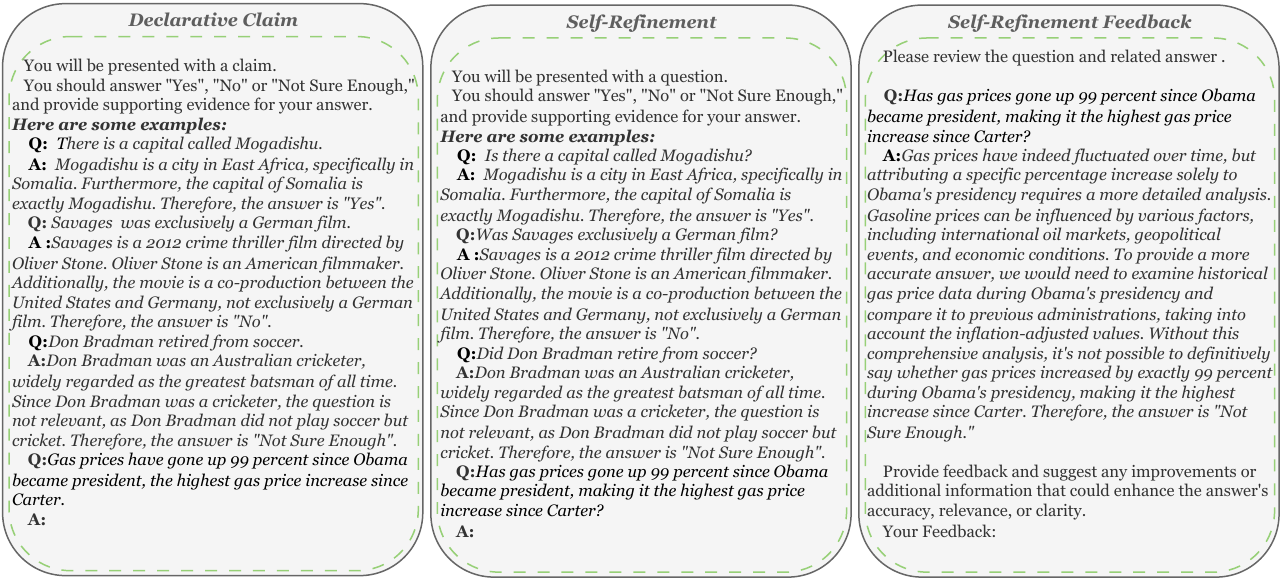}
   \caption{Prompts of self-refinement and declarative claim method.}
 \end{figure}

\newpage
\subsection{Case Study}\label{case-study}
\begin{figure}[h]
  \centering
  \includegraphics[scale=0.6]{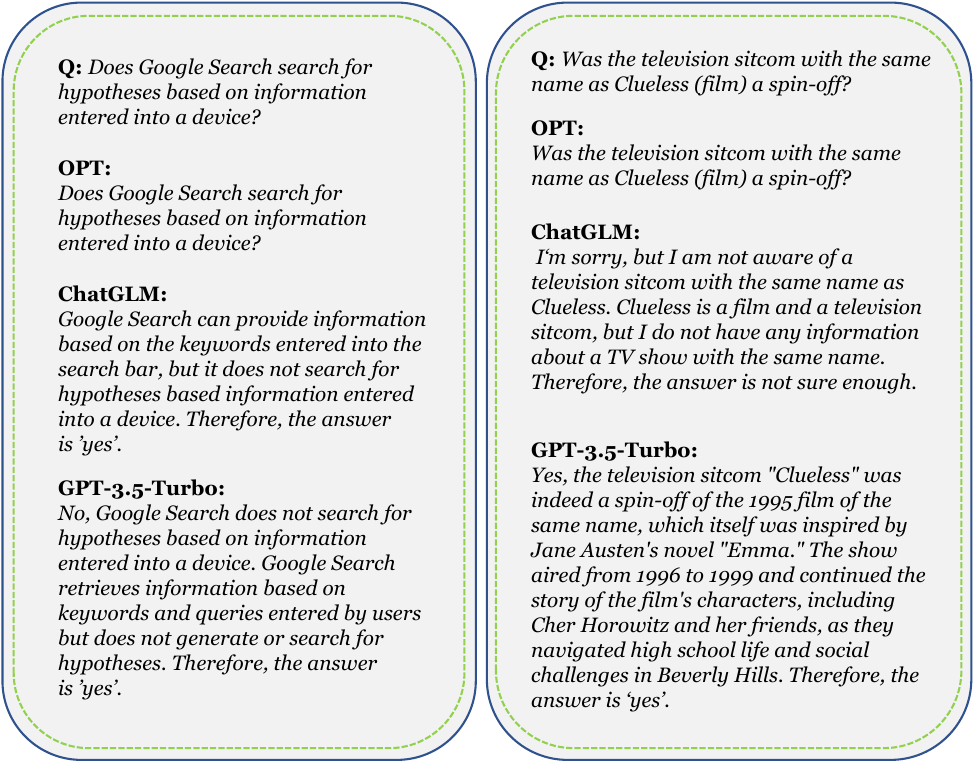}
  \caption{Answers to the same question from different LLMs in the zero-shot setting.}
  \label{fig:case-study}
\end{figure}

We have introduced an additional scenario for investigation, which occurs frequently in the output generated by the zero-shot prompt method. We conducted an experiment involving three models: OPT, ChatGLM, and GPT-3.5-Turbo. These models are presented with the same set of questions, and their responses are shown in Figure \ref{fig:case-study}. It is noteworthy that the OPT model, in both questions, reiterated the question itself without providing the corresponding answer.  It is essential to mention that the actual output of the OPT model repeats the problem until it reaches the maximum output length (controlled by the "max\_length" parameter), and we truncated the repeated portion.

The OPT model even declined to generate any content when presented with the zero-shot prompt, resulting in a significant number of empty responses in the statistical results. In the first question, both ChatGLM and GPT-3.5-Turbo provided correct answers. However, in the second question, when faced with more detailed information inquiries, ChatGLM failed to produce a correct response, while GPT-3.5-Turbo demonstrated proficient reasoning and provided accurate answers.

\end{document}